\documentclass[11pt]{article}

\usepackage[final]{acl}

\usepackage{times}
\usepackage{latexsym}
\usepackage{amsmath}
\usepackage{microtype}
\usepackage{inconsolata}
\usepackage{graphicx}

\usepackage{fontspec}
\usepackage{xeCJK}

\usepackage{booktabs}
\usepackage{multirow}
\usepackage{makecell}
\usepackage{tcolorbox}
\usepackage{xcolor}
\definecolor{translationcolor}{RGB}{0,102,204}
\definecolor{PaperGreen}{RGB}{0,128,0}
\definecolor{PaperRed}{RGB}{200,0,0}
\definecolor{teal}{RGB}{0,128,128}
\definecolor{gray}{RGB}{128,128,128}

\newcommand{\AppendixLayout}{%
  \setlength{\textfloatsep}{8pt}%
  \setlength{\floatsep}{6pt}%
  \setlength{\intextsep}{6pt}%
}

\title{CCiV: A Benchmark for Structure, Rhythm and Quality in LLM-Generated Chinese \textit{Ci} Poetry}

\author{
  Shangqing Zhao$^{1}$ \quad
  Yupei Ren$^{1,2}$ \quad
  Yuhao Zhou$^{1}$ \quad
  Xiaopeng Bai$^{2,3}$ \quad
  Man Lan$^{1,2}$\thanks{Corresponding author.}\\
  $^{1}$ School of Computer Science and Technology, East China Normal University, China\\
  $^{2}$ Shanghai Institute of Artificial Intelligence for Education, East China Normal University, China\\
  $^{3}$ Department of Chinese Language and Literature, East China Normal University, China\\
}

\begin{document}
\maketitle

\begin{abstract}
The generation of classical Chinese \textit{Ci} poetry, a form demanding a sophisticated blend of structural rigidity, rhythmic harmony, and artistic quality, poses a significant challenge for large language models (LLMs). To systematically evaluate and advance this capability, we introduce \textbf{C}hinese \textbf{Ci}pai \textbf{V}ariants (\textbf{CCiV}), a benchmark designed to assess LLM-generated \textit{Ci} poetry across these three dimensions: structure, rhythm, and quality. Our evaluation of 17 LLMs on 30 \textit{Cipai} reveals two critical phenomena: models frequently generate valid but unexpected historical variants of a poetic form, and adherence to tonal patterns is substantially harder than structural rules. We further show that form-aware prompting can improve structural and tonal control for stronger models, while potentially degrading weaker ones. Finally, we observe weak and inconsistent alignment between formal correctness and literary quality in our sample. CCiV highlights the need for variant-aware evaluation and more holistic constrained creative generation methods.
\end{abstract}


\section{Introduction}
\label{sec:intro}

\begin{figure}[t]
    \centering
    \includegraphics[width=\columnwidth]{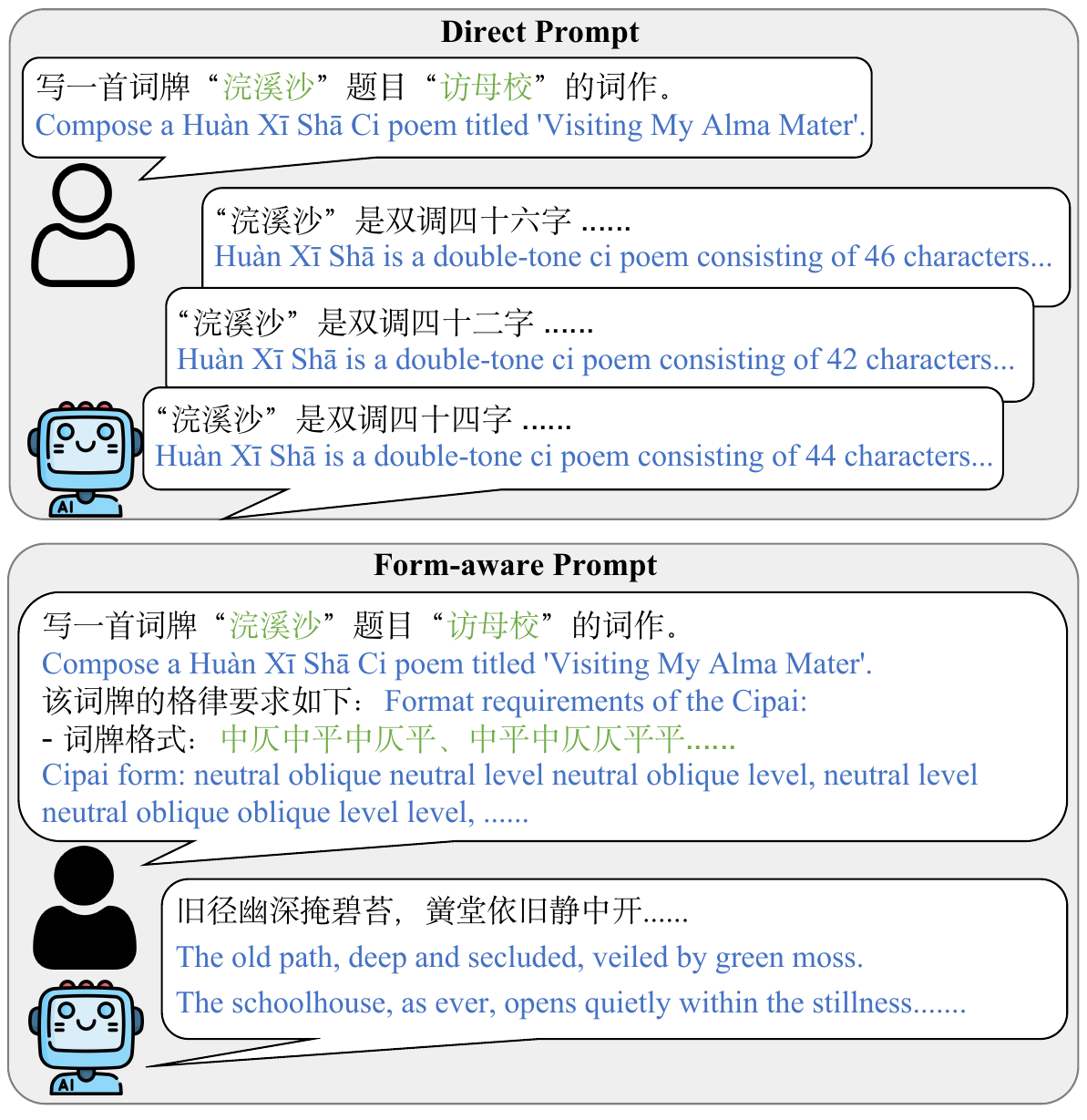}
    \caption{Prompting strategies for \textit{Ci} poetry generation: direct prompting versus form-aware prompting with explicit structural guidance.}
    \label{fig:prompting_strategy}
\end{figure}

\textit{Ci} (词) is a genre of Chinese poetry that flourished during the Song dynasty. Each \textit{Ci} poem must follow a specific \textit{Cipai} (词牌), or metric template, which specifies stanza count, lines per stanza, characters per line, tonal pattern (平仄, pīngzè, level and oblique tones), and rhyme scheme. Many \textit{Cipai} have multiple historical variants with different structural or tonal requirements, making \textit{Ci} a stringent testbed for controllable generation and evaluation. Figure~\ref{fig:prompting_strategy} illustrates the two prompting conditions used in our study.

Classical Chinese poetry generation has evolved significantly over time. Early approaches using template-based, statistical, and human-involved methods~\cite{li2018GeneratingClassicalChinese,zhipeng2019JiugeHumanMachineCollaborative} relied on limited keyword inputs and often produced rigid, semantically poor outputs~\cite{cao2024SurveyDeepLearning}. With the advancement of deep learning, recurrent neural networks (RNNs) and Transformer~\cite{vaswaniAttentionAllYou2017} models became mainstream approaches, offering better contextual understanding and overall consistency~\cite{wang2016ChinesePoetryGeneration,liu2020DeepPoetryChinese,shao2021SentimentStyleControllablea, li2021poetic}. More recently, systems like Yusheng~\cite{ma2023YuShengHumaninLoop} and Shisanbai~\cite{aichpoem2025} have demonstrated practical applications of Transformer-based generation. The emergence of large language models (LLMs) has further transformed this landscape, allowing for unrestricted user prompts and enabling better content control compared to traditional keyword-based systems~\cite{zou2025BIProZeroshotChinese, yu2024CharPoetChineseClassicala}.
In contrast, most research still focuses on \textit{Shi} forms such as regulated verse (\textit{律诗}) and quatrains (\textit{绝句}), which have less structural complexity than \textit{Ci}. Existing benchmarks include \textit{Ci} composition~\cite{cao2024WenMindComprehensiveBenchmark,zhao2025fuxibenchmarkevaluatinglanguage}, but they are limited in scale and typically ignore historical variants. As a result, the structural complexity of \textit{Cipai} forms and their variants has not been systematically evaluated.

To address this gap, we introduce \textbf{CCiV}, a benchmark that systematically evaluates LLM-generated Chinese \textit{Ci} poetry across structure, rhythm, and quality while explicitly accounting for historical variants. Our study addresses three research questions:

\textbf{RQ1:} How accurately can modern LLMs generate \textit{Ci} poetry that satisfies strict structural and tonal constraints?

\textbf{RQ2:} What are the common failure modes of LLMs in this task, particularly concerning semantic quality and adherence to form?

\textbf{RQ3:} Can explicit structural guidance in prompts improve \textit{Ci} poetry generation quality?

Our evaluation yields three findings: \textbf{(a)} LLMs struggle more with tonal constraints than structural requirements; \textbf{(b)} models often generate valid historical variants instead of standard forms, causing standard evaluation to underestimate capability; and \textbf{(c)} form-aware prompting improves control for stronger models but can hurt smaller ones. These findings motivate the benchmark design and analysis contributions below.

Our contributions are:
\begin{itemize}
\item \textbf{CCiV Benchmark}: A benchmark that evaluates \textit{Ci} poetry across structure, rhythm, and quality, using variant-aware evaluation.
\item \textbf{Large-Scale Evaluation}: Evaluation of 17 LLMs across 30 \textit{Cipai} forms, revealing variant generation and the mixed effects of form-aware prompting.
\item \textbf{Quality-Accuracy Analysis}: Analysis showing that models leading in structural/tonal metrics often do not produce the most aesthetically pleasing outputs.
\end{itemize}

\section{Methodology}
\label{sec:method}

We first describe the data collection process and experimental setup, then define the evaluation metrics used throughout our study.

\subsection{Data Collection and Preparation}

We constructed CCiV from two primary sources. First, we collected 49,270 \textit{Ci} poems from a public database\footnote{\url{https://github.com/yxcs/poems-db}} and identified the 30 most frequently occurring \textit{Cipai} forms, which collectively represent diverse structural complexity and historical prevalence. Second, for each \textit{Cipai}, we retrieved complete definition and variant information from a Chinese cultural knowledge graph\footnote{\url{https://cnkgraph.com/CiTune}}, including the standard form and all documented historical variants. For evaluation prompts, we randomly sample 10 titles from poems of the same \textit{Cipai} in the corpus, yielding 300 prompts in total. CCiV is evaluation-only; no model training or fine-tuning is performed. Pairing a large poem corpus with explicit template knowledge enables variant-aware evaluation under controlled prompts.

Table~\ref{tab:cipai_stat} summarizes key statistics for the 30 \textit{Cipai} forms selected for CCiV, revealing substantial variation in structural complexity: character counts range from 27 (e.g., \textit{望江南}, \textit{忆江南}) to 116 (e.g., \textit{金缕曲}, \textit{贺新郎}), while the number of structural variants per \textit{Cipai} ranges from 1 to 26. This diversity ensures CCiV evaluates models across a comprehensive spectrum of difficulty levels and supports difficulty-aware analysis in later sections. English descriptions of all \textit{Cipai} forms are provided in Appendix~\ref{app:cipai_details}.

\begin{table}[t]
    \caption{Statistics for the 30 \textit{Cipai} forms in CCiV. ``Works Count'' shows the number of historical poems in our dataset; percentage indicates relative frequency among all \textit{Ci} poems. ``Std. Chars'' is the character count for the standard form. ``Structure Variants'' is the number of documented historical variants with different structural patterns. ``Avg. Chars'' is the mean character count across all variants. Forms are ordered by standard character count (descending).}
    \label{tab:cipai_stat}
    \centering
    \resizebox{0.98\columnwidth}{!}{%
    \begin{tabular}{@{}lccccc@{}}
        \toprule
        \textbf{Cipai} & \textbf{Works Count}  & \textbf{Std. Chars}  & \makecell{\textbf{Structure}\\\textbf{Variants}}  & \textbf{Avg. Chars} \\ \midrule
        C1.金缕曲      & 580 (2.02\%)     & 116        & 11         & 115.64      \\
        C2.贺新郎      & 409 (1.43\%)     & 116        & 11         & 115.64      \\
        C3.摸鱼儿      & 329 (1.15\%)     & 116        & 6          & 115.50      \\
        C4.沁园春      & 651 (2.27\%)     & 114        & 8          & 114.25      \\
        C5.水龙吟      & 543 (1.89\%)     & 102        & 26         & 102.38      \\
        C6.念奴娇      & 618 (2.16\%)     & 100        & 11         & 100.27      \\
        C7.水调歌头     & 743 (2.59\%)     & 95         & 7          & 95.57      \\
        C8.满庭芳      & 610 (2.13\%)     & 95         & 7          & 95.29      \\
        C9.满江红      & 950 (3.32\%)     & 93         & 13         & 92.85      \\
        C10.渔家傲      & 292 (1.02\%)     & 62         & 2          & 64.00      \\
        C11.蝶恋花      & 1029 (3.59\%)    & 60         & 1          & 60.00      \\
        C12.临江仙      & 971 (3.39\%)     & 58         & 8          & 58.38      \\
        C13.踏莎行      & 556 (1.94\%)     & 58         & 3          & 62.67      \\
        C14.鹊桥仙      & 209 (0.73\%)     & 56         & 5          & 63.40      \\
        C15.虞美人      & 686 (2.39\%)     & 56         & 3          & 56.67      \\
        C16.鹧鸪天      & 1414 (4.93\%)    & 55         & 1          & 55.00      \\
        C17.西江月      & 600 (2.09\%)     & 50         & 3          & 52.33      \\
        C18.清平乐      & 770 (2.69\%)     & 46         & 2          & 46.00      \\
        C19.采桑子      & 378 (1.32\%)     & 44         & 3          & 48.67      \\
        C20.卜算子      & 393 (1.37\%)     & 44         & 6          & 45.17      \\
        C21.减字木兰花    & 561 (1.96\%)     & 44         & 1          & 44.00      \\
        C22.菩萨蛮      & 1268 (4.43\%)    & 44         & 1          & 44.00      \\
        C23.浣溪沙      & 1692 (5.9\%)    & 42         & 3          & 44.00      \\
        C24.点绛唇      & 567 (1.98\%)     & 41         & 3          & 41.67      \\
        C25.江城子      & 219 (0.76\%)     & 35         & 5          & 42.80      \\
        C26.如梦令      & 442 (1.54\%)     & 33         & 3          & 44.00      \\
        C27.浪淘沙      & 548 (1.91\%)     & 28         & 1          & 28.00      \\
        C28.南乡子      & 435 (1.52\%)     & 27         & 9          & 43.89      \\
        C29.望江南      & 532 (1.86\%)     & 27         & 3          & 46.67      \\
        C30.忆江南      & 322 (1.12\%)     & 27         & 3          & 46.67      \\
        \bottomrule
        \end{tabular}%
    }
\end{table}

\subsection{Experimental Setup}

\subsubsection{Model Selection}

We selected 17 LLMs spanning GPT~\cite{openai2023GPT4TechnicalReport}, Qwen~\cite{yang2024qwen2technicalreport, qwen2025qwen25technicalreport} (including QwQ and Ds-Qwen), GLM~\cite{glm2024chatglm}, Llama~\cite{grattafiori2024Llama3Herd}, InternLM~\cite{cai2024internlm2}, and Deepseek-R1~\cite{guo2025DeepSeekR1IncentivizesReasoning} to ensure comprehensive coverage of architectures, scales, and training approaches. For open-weight models, sizes range from 1.5B to 72B parameters; API model sizes are undisclosed. All models are evaluated under the two prompting conditions described next.

\subsubsection{Prompting Conditions}

We designed two prompting conditions to assess different aspects of model capability:

\noindent \textbf{Direct Prompt}: The model receives only the \textit{Cipai} name and theme. This condition tests the model's internal knowledge of \textit{Ci} forms acquired during pre-training. To account for the variant generation phenomenon, we evaluate outputs against both the standard form and all known historical variants; a match with any variant is considered correct. This dual evaluation strategy reveals the extent to which models have learned the distribution of historical forms.

\noindent \textbf{Form-aware Prompt}: The prompt explicitly provides the complete structural and tonal template for the standard form. This instruction-only format specification provides exact line lengths and tonal requirements without semantic examples that could bias content. This condition tests the model's ability to follow explicit structural constraints while generating creative content. Together, the two prompting conditions isolate the effect of explicit structural guidance.

\subsubsection{Decoding and Sampling}

For each model, \textit{Cipai}, and theme, we sample three outputs using temperature $T=0.7$ and top-$p=0.95$. For local models, we fix decoding parameters and random seeds when supported to reduce variance; for API models, decoding stochasticity cannot be fully controlled, so we report averages across three runs. All reported metrics are averages over these three samples, unless otherwise stated.

Detailed prompt templates and examples are provided in Appendix~\ref{app:exp_details}.

\begin{figure}[t]
    \centering
    \includegraphics[width=0.85\columnwidth]{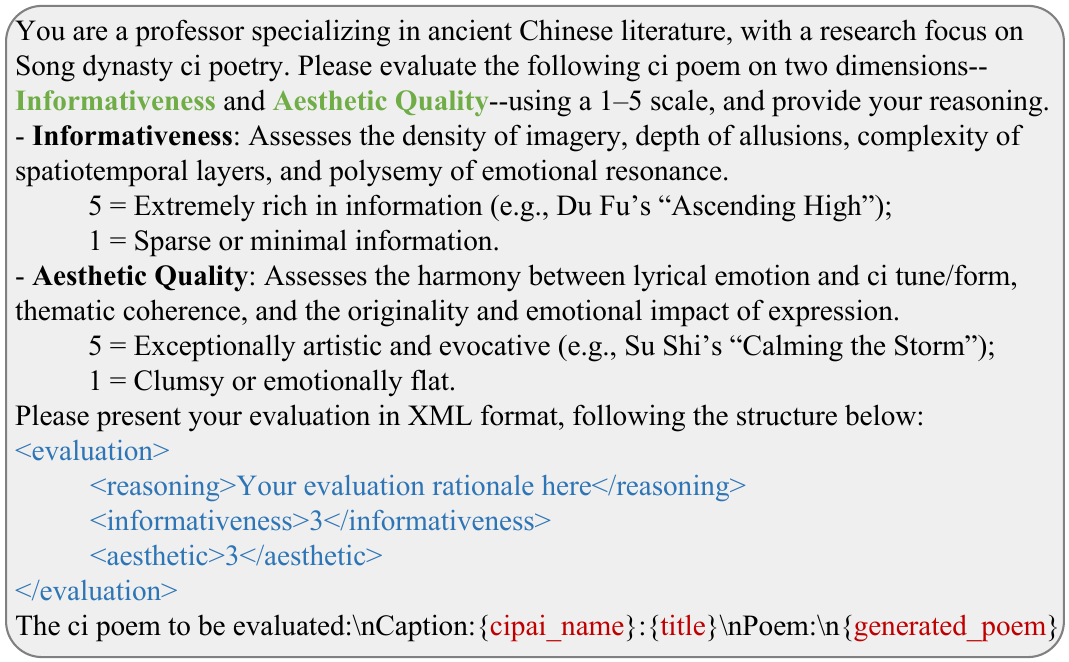}
    \caption{Prompt template for evaluating Informativeness and Aesthetic metrics.}
    \label{fig:eval_prompt}
\end{figure}

\subsection{Evaluation Metrics}

\noindent We assess three dimensions in order: structure, tone, and semantic quality.

\textbf{Structural Accuracy}: A \textit{Cipai} prescribes a poem's structure, including stanza count, line count, and characters per line. A generated poem is considered structurally accurate if its total line count and per-line character counts exactly match a target template (standard or variant). We normalize outputs by removing non-Chinese characters via regex, replacing separators with line breaks, and trimming whitespace before line segmentation. The score is binary (1 for exact match, 0 otherwise), as defined in Eq.~(1). For variant-aware evaluation, we consider the generation correct if it matches any valid variant of the requested \textit{Cipai}, as shown in Eq.~(2). We report both standard and variant-aware scores; the variant-aware metric captures historically valid alternatives rather than serving as a cross-\textit{Cipai} fairness metric.
\noindent If a generation is structurally incorrect, we assign a structural score of 0 and set its tonal score to 0.
\begin{align}
\text{Acc}(g, t) =
\begin{cases}
1 & |P| = |S| \land |p_j| = |s_j|, \\
  &\ \forall j \in \{1,\dots,|S|\} \\
0 & \text{otherwise}
\end{cases}
\end{align}
where $g$ is the LLM-generated text, $t$ is a \textit{Cipai} form, $S$ is the sequence of sentences in $t$, and $P$ is the sequence of parsed sentences from $g$.
\begin{equation}
\text{Acc}_{\text{variants}}(g, \mathcal{T}) = \max_{t \in \mathcal{T}} \text{Acc}(g, t)
\end{equation}
where $\mathcal{T}$ is the set of all valid templates for a given \textit{Cipai}.

\noindent \textbf{Tonal Accuracy}: We calculate the percentage of characters matching the required tonal pattern, defined in Eq.~(3). Character tones are identified using the 中华新韵 (Zhonghua Xinyun) standard.
\begin{equation}
\text{Acc}_{\text{tonal}}(g, t) = \frac{1}{N} \sum_{i=1}^{N} \delta(\tau(C(g)_i), \mathcal{P}(t)_i)
\end{equation}
where $C(g)$ is the sequence of N characters in $g$, $\mathcal{P}(t)$ is the corresponding sequence of required tones, and $\delta$ is the tonal matching function. As with structural accuracy, we compute variant-aware Tonal Accuracy by taking the maximum score across all valid variants.

\noindent \textbf{Quality Evaluation}: Finally, to assess semantic and artistic quality, we employed an LLM-as-a-judge approach~\cite{zheng2023judging}. Given the large volume of generated poems, we evaluated results from four representative models. The evaluation focused on two criteria, rated on a 1-to-5 scale: \textbf{Informativeness}, measuring the substance of the poem's content, and \textbf{Aesthetic}, assessing its artistic merit and coherence with the given theme. We report mean scores across the three judges and the three sampled generations per prompt, and compute variance and 95\% confidence intervals across all judged samples (normal approximation, mean $\pm$ 1.96\,SE; not shown due to space). This qualitative assessment complements our automated metrics by providing insights into creative capabilities beyond formal accuracy.

\section{Results and Analysis}
\label{sec:results}

Having established our evaluation framework, we now present experimental results across 17 LLMs. We start with direct prompting and the variant generation phenomenon, then analyze tonal adherence and the effects of form-aware prompting, and finally connect formal accuracy to LLM-judged quality.

\begin{table*}[t]
    \caption{Model performance (\%) across Direct and Form-aware prompting conditions. For Direct Prompt, we evaluate against both standard (Std.) and all historical variant (Var.) forms. For Form-aware, models are guided to produce a specific standard form. Compared with Direct Prompt Std., changes in \textcolor{teal}{teal} indicate improvement, \textcolor{red}{red} indicate decline, \textcolor{gray}{gray} indicate no change.}
    \label{tab:main_results}
    \centering
    \small
    \resizebox{0.95\textwidth}{!}{%
    \begin{tabular}{l|cccc|cc}
    \toprule
    \multicolumn{1}{c|}{\multirow{3}{*}{\textbf{Model}}} &
      \multicolumn{4}{c|}{\textbf{Direct Prompt}} &
      \multicolumn{2}{c}{\textbf{Form-aware Prompt}} \\ \cline{2-7}
    \multicolumn{1}{c|}{} &
      \multicolumn{2}{c}{\textbf{Structural Acc.}} &
      \multicolumn{2}{c|}{\textbf{Tonal Acc.}} &
      \textbf{Structural Acc.} &
      \textbf{Tonal Acc.} \\
    \multicolumn{1}{c|}{} &
      \textbf{Std.} &
      \textbf{Var.} &
      \textbf{Std.} &
      \textbf{Var.} &
      \textbf{Std.} &
      \textbf{Std.} \\ \midrule
    GPT-4o-mini         & 0.02  & 0.02\,\textcolor{gray}{\footnotesize{0.00}}  & 0.02  & 0.02\,\textcolor{gray}{\footnotesize{0.00}}  & 7.33\,\textcolor{teal}{\footnotesize{+7.31}}  & 5.10\,\textcolor{teal}{\footnotesize{+5.08}}  \\
    GPT-4o              & 55.33 & 68.00\,\textcolor{teal}{\footnotesize{+12.67}} & 49.36 & 61.94\,\textcolor{teal}{\footnotesize{+12.58}} & 63.00\,\textcolor{teal}{\footnotesize{+7.67}} & 55.10\,\textcolor{teal}{\footnotesize{+5.74}} \\
    Qwen-max            & 39.00 & 50.67\,\textcolor{teal}{\footnotesize{+11.67}} & 37.05 & 48.42\,\textcolor{teal}{\footnotesize{+11.37}} & \textbf{78.67}\,\textcolor{teal}{\footnotesize{+39.67}} & \textbf{74.83}\,\textcolor{teal}{\footnotesize{+37.78}} \\
    GLM-4-plus          & 42.00 & 51.00\,\textcolor{teal}{\footnotesize{+9.00}} & 37.45 & 45.99\,\textcolor{teal}{\footnotesize{+8.54}} & 70.00\,\textcolor{teal}{\footnotesize{+28.00}} & 61.63\,\textcolor{teal}{\footnotesize{+24.18}} \\
    Deepseek-R1         & 56.67 & 67.33\,\textcolor{teal}{\footnotesize{+10.66}} & 53.13 & 63.73\,\textcolor{teal}{\footnotesize{+10.60}} & 66.67\,\textcolor{teal}{\footnotesize{+10.00}} & 63.18\,\textcolor{teal}{\footnotesize{+10.05}} \\
    Ds-Qwen-32b         & 28.00 & 38.00\,\textcolor{teal}{\footnotesize{+10.00}} & 24.92 & 34.58\,\textcolor{teal}{\footnotesize{+9.66}} & 28.33\,\textcolor{teal}{\footnotesize{+0.33}} & 24.06\,\textcolor{red}{\footnotesize{-0.86}} \\
    QwQ-32b             & 12.00 & 14.67\,\textcolor{teal}{\footnotesize{+2.67}} & 11.29 & 13.98\,\textcolor{teal}{\footnotesize{+2.69}} & 42.67\,\textcolor{teal}{\footnotesize{+30.67}} & 38.56\,\textcolor{teal}{\footnotesize{+27.27}} \\
    Internlm2.5-20b     & 5.33  & 5.67\,\textcolor{teal}{\footnotesize{+0.34}}  & 4.35  & 4.79\,\textcolor{teal}{\footnotesize{+0.44}}  & 1.00\,\textcolor{red}{\footnotesize{-4.33}}  & 0.93\,\textcolor{red}{\footnotesize{-3.42}}  \\
    Internlm2.5-7b      & 18.67 & 20.00\,\textcolor{teal}{\footnotesize{+1.33}} & 15.55 & 17.17\,\textcolor{teal}{\footnotesize{+1.62}} & 14.33\,\textcolor{red}{\footnotesize{-4.34}} & 11.88\,\textcolor{red}{\footnotesize{-3.67}} \\
    GLM4-9b             & 6.33  & 6.67\,\textcolor{teal}{\footnotesize{+0.34}}  & 4.64  & 4.93\,\textcolor{teal}{\footnotesize{+0.29}}  & 0.33\,\textcolor{red}{\footnotesize{-6.00}}  & 0.27\,\textcolor{red}{\footnotesize{-4.37}}  \\
    Llama3.1-8b         & 0.33  & 0.33\,\textcolor{gray}{\footnotesize{0.00}}  & 0.22  & 0.22\,\textcolor{gray}{\footnotesize{0.00}}  & 0.00\,\textcolor{red}{\footnotesize{-0.33}}   & 0.00\,\textcolor{red}{\footnotesize{-0.22}}  \\
    Llama3.1-8b-chinese & 0.67  & 0.67\,\textcolor{gray}{\footnotesize{0.00}}  & 0.40  & 0.40\,\textcolor{gray}{\footnotesize{0.00}}  & 0.00\,\textcolor{red}{\footnotesize{-0.67}}   & 0.00\,\textcolor{red}{\footnotesize{-0.67}}     \\
    Qwen2.5-1.5b        & 20.67 & 26.33\,\textcolor{teal}{\footnotesize{+5.66}} & 18.29 & 23.66\,\textcolor{teal}{\footnotesize{+5.37}} & 4.00\,\textcolor{red}{\footnotesize{-16.67}} & 3.58\,\textcolor{red}{\footnotesize{-14.71}} \\
    Qwen2.5-3b          & 32.67 & 38.00\,\textcolor{teal}{\footnotesize{+5.33}} & 7.85  & 12.37\,\textcolor{teal}{\footnotesize{+4.52}} & 15.00\,\textcolor{red}{\footnotesize{-17.67}} & 13.70\,\textcolor{teal}{\footnotesize{+5.85}} \\
    Qwen2.5-7b          & 46.67 & 58.33\,\textcolor{teal}{\footnotesize{+11.66}} & 1.05 & 1.18\,\textcolor{teal}{\footnotesize{+0.13}} & 30.00\,\textcolor{red}{\footnotesize{-16.67}} & 27.42\,\textcolor{teal}{\footnotesize{+26.37}} \\
    Qwen2.5-14b         & 43.33 & 55.67\,\textcolor{teal}{\footnotesize{+12.34}} & 40.96 & 53.01\,\textcolor{teal}{\footnotesize{+12.05}} & 49.00\,\textcolor{teal}{\footnotesize{+5.67}} & 45.09\,\textcolor{teal}{\footnotesize{+4.13}} \\
    Qwen2.5-72b         & \textbf{61.67} & \textbf{80.33}\,\textcolor{teal}{\footnotesize{+18.66}} & \textbf{57.00} & \textbf{75.98}\,\textcolor{teal}{\footnotesize{+18.98}} & 70.34\,\textcolor{teal}{\footnotesize{+8.67}} & 65.22\,\textcolor{teal}{\footnotesize{+3.55}} \\  \bottomrule
    \end{tabular}%
    }
\end{table*}

\subsection{Baseline Performance and Variant Generation}

Table~\ref{tab:main_results} shows performance under both conditions. For Direct Prompt, we report accuracy against the standard form and all historical variants.

Under direct prompting, most models score low against the standard form alone. Only three models exceed 55\% structural accuracy: Qwen2.5-72b (61.67\%), Deepseek-R1 (56.67\%), and GPT-4o (55.33\%). When evaluated against all variants, accuracy increases. Qwen2.5-72b improves by 18.66 points, GPT-4o by 12.67 points, and Qwen-max by 11.67 points. Many apparent errors are actually correct generations of alternative forms, providing evidence for the variant generation phenomenon. We next inspect per-\textit{Cipai} patterns to see where these gains concentrate.

\begin{figure}[t]
    \centering
    \includegraphics[width=\columnwidth]{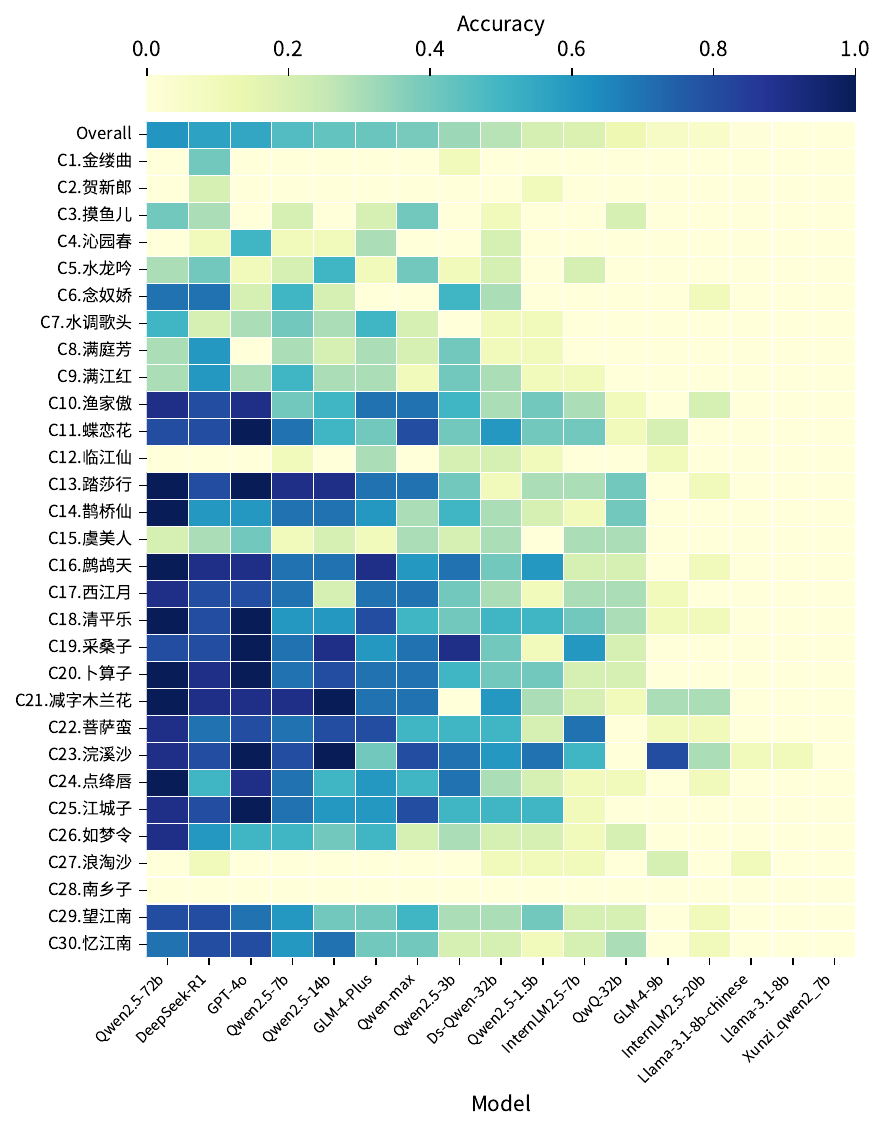}
    \caption{Heatmap of structural accuracy against standard forms under direct prompting. Each row represents a model (ordered by overall performance), and each column represents a \textit{Cipai} form (ordered by character count, right-to-left: 116 to 27). Darker cells indicate higher accuracy. Two patterns emerge: (1) better performance on shorter forms (left) than longer forms (right), and (2) stark performance gaps between model families. Anomalous low-accuracy columns (e.g., C27, C28) reflect the variant generation phenomenon.}
    \label{fig:direct_prompt_heatmap}
\end{figure}

The heatmap in Figure~\ref{fig:direct_prompt_heatmap} provides a granular view of model performance across \textit{Cipai} forms. Two clear patterns emerge: (1) a moderate negative correlation between character count and accuracy—models perform substantially better on shorter forms (bottom-left, e.g., C29, C30: 27 chars) than longer ones (top-right, e.g., C1, C2: 116 chars). Quantitatively, across 30 \textit{Cipai}, average structural accuracy under direct prompting correlates negatively with standard character count (Spearman $\rho=-0.36$, $p=0.050$, averaged across 17 models); and (2) a massive performance gap between model families—top models like Qwen2.5-72b and Deepseek-R1 achieve high accuracy across most forms, while smaller models like Llama3.1-8b consistently score near zero.

Notably, the heatmap reveals anomalous low scores for specific \textit{Cipai} such as C27 (\textit{浪淘沙}), C28 (\textit{南乡子}), and C15 (\textit{虞美人}), where nearly all models perform poorly despite moderate character counts. Appendix~\ref{app:var_eval} demonstrates that these anomalies are direct consequences of variant generation: models consistently generate popular historical variants (e.g., \textit{浪淘沙令} instead of \textit{浪淘沙}) rather than the standard forms used for evaluation, likely reflecting the distribution of these forms in pre-training data.

\subsection{The Challenge of Tonal Adherence}

Tonal patterns are as critical as structure in \textit{Ci} poetry, requiring precise character-level tone assignments. Table~\ref{tab:main_results} shows this is harder than structural compliance. Deepseek-R1 achieves 53.13\% tonal accuracy, 3.54 points below its structural accuracy (56.67\%). The gap is larger for other models: GPT-4o shows a 5.97-point gap (49.36\% vs 55.33\%), and Qwen2.5-72b shows a 4.67-point gap (57.00\% vs 61.67\%). LLMs can learn sentence-level structural patterns, but character-level tonal constraints are more difficult.

Variant-aware evaluation improves tonal accuracy, but scores still lag behind structural accuracy. Qwen2.5-72b achieves the best variant-aware tonal accuracy (75.98\%), which is 4.35 points below its structural accuracy (80.33\%). This motivates testing whether explicit templates can further improve tonal control.

\subsection{Improving Control with Form-Aware Prompting}

We investigated whether explicit structural guidance could enforce the standard form. Table~\ref{tab:main_results} shows that form-aware prompting helps some models but hurts others.

For models with strong instruction-following, the improvements are large. Qwen-max shows the biggest gain: structural accuracy increases by 39.67 points (39.00\% to 78.67\%), and tonal accuracy improves by 37.78 points (37.05\% to 74.83\%). GLM-4-plus gains 28.00 points in structural accuracy and 24.18 points in tonal accuracy. GPT-4o gains 7.67 and 5.74 points respectively. These results show that capable models can follow precise instructions to achieve specific structures.

For smaller models, form-aware prompting hurts performance. Qwen2.5-7b's structural accuracy drops by 16.67 points (46.67\% to 30.00\%). Qwen2.5-3b's tonal accuracy improves only 5.85 points despite its structural degradation of 17.67 points. The detailed template seems to overwhelm these models' capacity.

The largest model (Qwen2.5-72b) shows smaller gains (8.67 points structural, 3.55 points tonal) than Qwen-max, possibly because its pre-training already provides strong representations of poetic forms. We therefore examine whether improved formal control translates into higher perceived quality.

\subsection{Qualitative Analysis}

Table~\ref{tab:human_eval} shows LLM-as-a-judge scores. Three top-tier LLMs (Kimi-K2~\cite{team2025kimi}, Doubao-seed-1.6~\cite{seed2025SeedThinkingv15AdvancingSuperb}, and Gemini-2.5-pro~\cite{comanici2025gemini}) evaluated Informativeness and Aesthetic on a 1-5 scale using the prompt in Figure~\ref{fig:eval_prompt}.

Formal accuracy does not reliably predict artistic quality in our sample. Deepseek-R1 ranks third in structural accuracy (66.67\%) and third in tonal accuracy (63.18\%), but achieves the highest Informativeness (3.28) and Aesthetic (3.37) scores. Qwen-max leads in formal metrics (78.67\% structural, 74.83\% tonal) but ranks third in Informativeness (2.82) and second in Aesthetic (2.99). Strict adherence to form may constrain creative expression.

All models score below 3.5 on average. LLMs can achieve high formal correctness, but generating poetry with literary merit remains difficult. This gap motivates the broader implications discussed next.

\begin{table}[t]
\caption{Quality evaluation of Form-aware Prompt results on four representative models. Structural and Tonal Accuracy are automated metrics; Informativeness and Aesthetic scores are from LLM-as-a-judge evaluation (three top-tier models: Kimi-K2, Doubao-seed-1.6, Gemini-2.5-pro) on a 1-5 scale. Deepseek-R1 achieves the highest artistic quality despite not leading in formal metrics.}
\label{tab:human_eval}
\centering
\resizebox{\columnwidth}{!}{%
\begin{tabular}{@{}lcccc@{}}
\toprule
\textbf{Model} & \makecell{\textbf{Structural}\\\textbf{Acc. (\%)}} & \makecell{\textbf{Tonal}\\\textbf{Acc. (\%)}} & \makecell{\textbf{Informativeness}\\\textbf{(1-5)}} & \makecell{\textbf{Aesthetic}\\\textbf{(1-5)}} \\ \midrule
GPT-4o & 63.00 & 55.10 & 2.68 & 2.85\\
Qwen-max & 78.67 & 74.83 & 2.82 & 2.99 \\
Deepseek-R1 & 66.67 & 63.18 & 3.28  & 3.37 \\
Qwen2.5-72b & 70.34 & 65.22 & 2.90 & 3.06 \\
\bottomrule
\end{tabular}%
}
\end{table}

\section{Discussion}
\label{sec:discussion}

Our findings reveal how LLMs handle rule-based creative tasks, with implications for evaluation and development. We summarize three implications below.

\noindent \textbf{The ``Variant Generation'' Phenomenon}: LLMs appear to learn a distribution of historical forms rather than a single canonical rule. Without explicit guidance, they sample from this distribution, producing outputs that are often valid but different from user expectations. This motivates variant-aware evaluation in constrained generation.

\noindent \textbf{The Structural-Tonal Gap}: Structural rules are discrete and explicit, while tonal constraints require fine-grained character-level phonetic control. This likely contributes to the persistent performance gap between structural and tonal accuracy across models.

\noindent \textbf{Form-Aware Prompting as a Double-Edged Sword}: Form-aware prompting is highly model-dependent~\cite{kaplan2020scalinglawsneurallanguage}. Larger models generally benefit from explicit templates, while smaller models may be overwhelmed by prompt complexity.

\section{Conclusion}
\label{sec:conclusion}

We introduced CCiV, a benchmark for evaluating LLM-generated Chinese \textit{Ci} poetry across structure, rhythm, and quality. Evaluating 17 LLMs across 30 \textit{Cipai}, we found that variant generation and tonal adherence remain core challenges. We also showed that form-aware prompting can improve control for capable models but may degrade weaker ones.

Our qualitative results further indicate weak alignment between formal correctness and artistic quality in this setting. Future work should improve explicit rhythm control and explore prompting strategies that better balance formal constraints and creativity.

\bibliography{strings}

\clearpage
\appendix
\AppendixLayout

\section{Experimental Details}
\label{app:exp_details}

This appendix provides prompt templates and supplementary analyses. Evaluation metrics are defined in Section~\ref{sec:method} and are not repeated here.

For reproducibility, we use temperature $T=0.7$ and top-$p=0.95$, and sample three generations per model--\textit{Cipai}--theme triple. Metrics in the main paper are averages across these samples. If a generation is structurally incorrect, both structural and tonal scores are set to 0. We do not evaluate rhyme positions or whether upper and lower stanzas share the same rhyme. LLM-as-a-judge scores are reported as the mean across three judges and three sampled generations; we compute variance and 95\% confidence intervals across all judged samples (normal approximation, mean $\pm$ 1.96\,SE). All prompts, templates, system prompts, and evaluation scripts are released with the benchmark at \url{https://github.com/cubenlp/CCiV} . 

\subsection{Prompt Examples}
\label{sec:example_prompts}

\noindent The following templates are the exact prompts used in evaluation. Figure~\ref{fig:zero_prompt} shows the direct (zero-shot) prompt without structural guidance, and Figure~\ref{fig:form_aware_prompt} shows the form-aware prompt with explicit template constraints. English translations and example outputs in red are for reader illustration only and are not included in the actual prompts. The schematic illustration in Figure~\ref{fig:prompting_strategy} in the main text is an example for explanation only.

\begin{figure*}[p]
\begin{tcolorbox}[width=\textwidth, colframe=blue!50!black, colback=blue!5!white, sharp corners, fontupper=\scriptsize]
    \textbf{Instruction:} \\按照提供的词牌名和题目写一首词, 要求严格遵守词牌的格律。\\
    \textcolor{translationcolor}{Write a \textit{Ci} poem according to the provided \textit{Cipai} name and title, strictly following the metrical rules of the \textit{Cipai}.} \\
    \textbf{Input:}\\
    词牌: \textcolor{PaperGreen}{望江南} \; \\
     \textcolor{translationcolor}{\textit{Cipai}: Gazing at the Yangtze's South} \\
    题目: \textcolor{PaperGreen}{红桥春游词} \; \\
     \textcolor{translationcolor}{Title: A Spring Outing at Hong Bridge} \\
    \textbf{Output:}\\
    \textcolor{PaperRed}{江南好, 修袅袅溪光。杨柳风微莺嘴滑, 桃花漾足鸭头香。几处洗衣裳。} \\
    \textcolor{translationcolor}{The South of Yangtze is lovely, \\
    I seize the creek's gleam for the spring ritual. \\
    In gentle breeze willows sway, while warblers' tongues slip sweet songs, \\
    Peach blossoms flood the scene with fragrance thick as ducks' heads, \\
    Here and there, women wash their clothes. }
\end{tcolorbox}
\caption{Direct prompt example (zero-shot). No additional structural guidance is provided. Text in green is example variable and text in red is example output.}
\label{fig:zero_prompt}
\end{figure*}

\begin{figure*}[p]
    \begin{tcolorbox}[width=\textwidth, colframe=blue!50!black, colback=blue!5!white, sharp corners, fontupper=\scriptsize]
        \textbf{Instruction:} \\请根据词牌``\textcolor{PaperGreen}{望江南}''创作一首词，主题为``\textcolor{PaperGreen}{红桥春游词}''。\\
        \textcolor{translationcolor}{Write a \textit{Ci} poem according to the provided \textit{Cipai} ``Gazing at the Yangtze's South'' and theme ``A Spring Outing at Hong Bridge''.} \\
        \textbf{Input:}\\
        该词牌的格律要求如下：\textcolor{translationcolor}{\ Format requirements of the \textit{Cipai}:} \\
         - 分句结构：\textcolor{PaperGreen}{平中仄、中仄仄平平、中仄中平平仄仄、中平中仄仄平平、中仄仄平平}\quad \\
         \textcolor{translationcolor}{sentence structure: level neutral oblique, neutral oblique oblique level level, neutral oblique neutral level level oblique oblique, neutral level neutral oblique oblique level level, neutral oblique oblique level level.} \\
        请直接输出词作，不需要解释。\\ \textcolor{translationcolor}{Please output the poem directly, without any explanation.} \\
        \textbf{Output:}\\
        \textcolor{PaperRed}{江南好，修袅袅溪光。杨柳风微莺嘴滑，桃花漾足鸭头香。几处洗衣裳。}\\
        \textcolor{translationcolor}{The South of Yangtze is lovely, \\
        I seize the creek's gleam for the spring ritual. \\
        In gentle breeze willows sway, while warblers' tongues slip sweet songs, \\
        Peach blossoms flood the scene with fragrance thick as ducks' heads, \\
        Here and there, women wash their clothes. }
    \end{tcolorbox}
    \caption{Form-aware prompt example. Explicit structural guidance is provided. Text in green is example variable and text in red is example output.}
    \label{fig:form_aware_prompt}
\end{figure*}

\section{Generation Examples and Additional Analyses}
\label{app:var_eval}

\begin{figure}[t]
    \centering
    \includegraphics[width=\columnwidth]{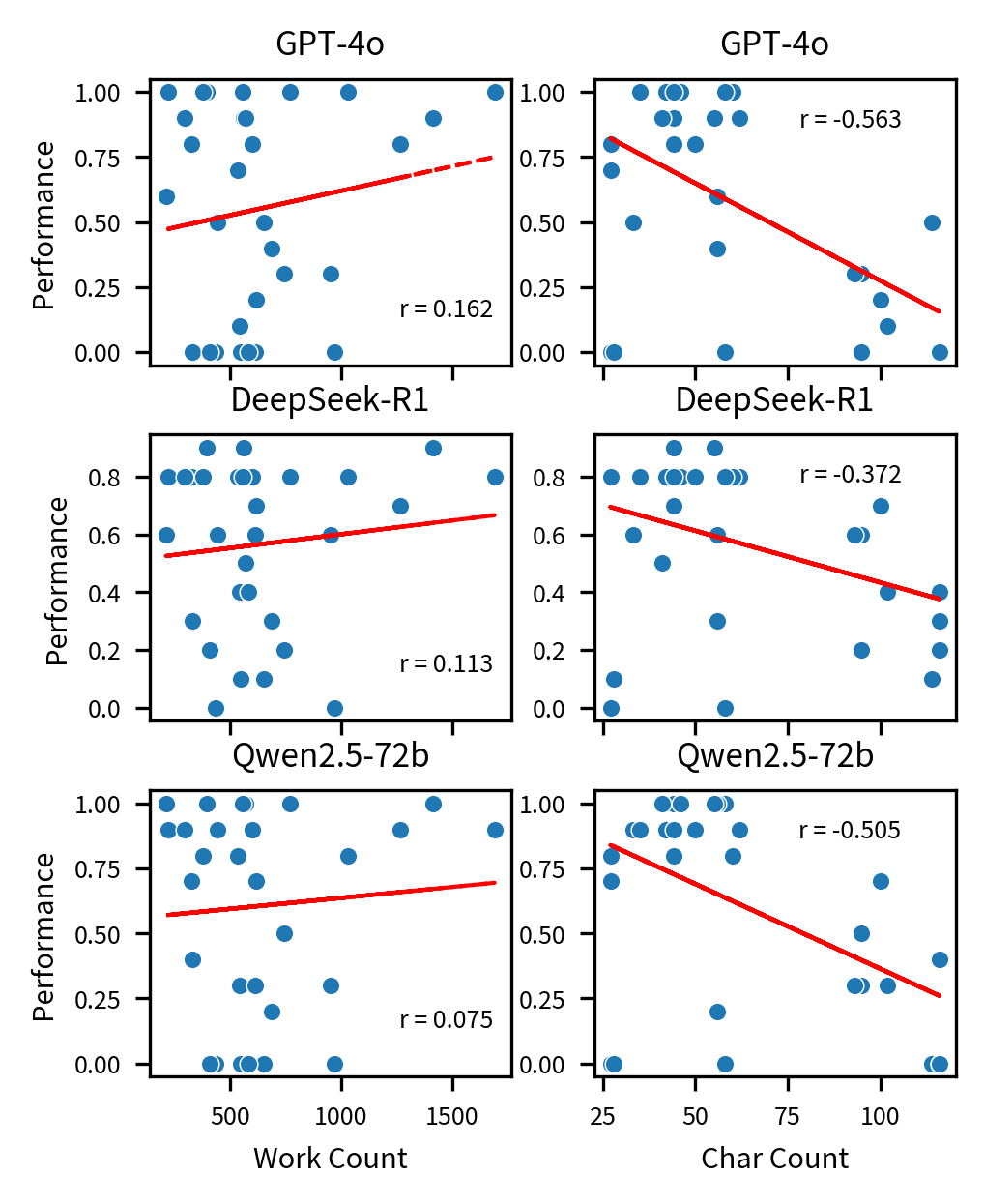}
    \caption{Performance of top models against number of historical works and character count for each \textit{Cipai} form.}
    \label{fig:top_models_scatter}
\end{figure}

\begin{figure}[t]
    \centering
    \includegraphics[width=\columnwidth]{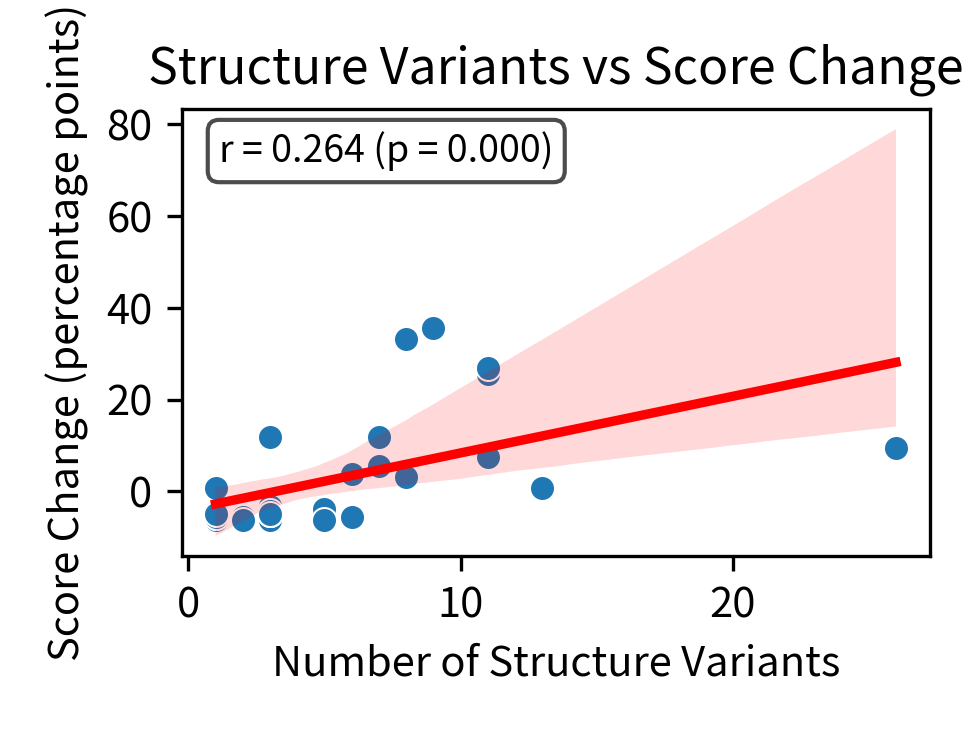}
    \caption{Average score change for each \textit{Cipai} correlates with number of variants.}
    \label{fig:score_change_variants_num}
\end{figure}

This section supplements Section~\ref{sec:results} with additional visual analyses. Figure~\ref{fig:top_models_scatter} relates performance to \textit{Cipai} statistics, and Figure~\ref{fig:score_change_variants_num} links variant-aware improvements to the number of historical variants.

\begin{figure*}[p]
    \centering
    \includegraphics[width=\textwidth]{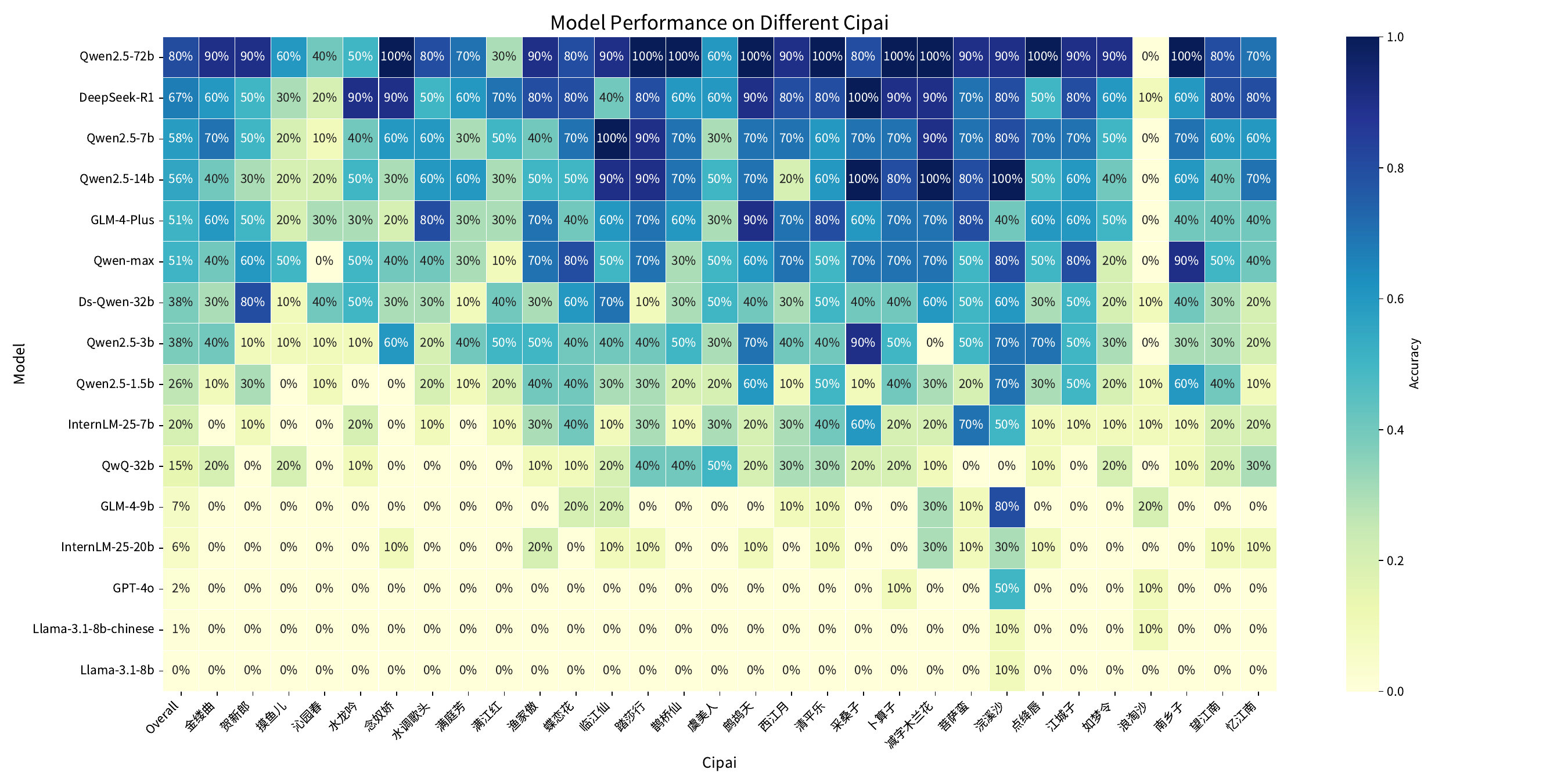}
    \caption{\textbf{Variants} setting: Heatmap of Model Performance on Different Cipai. Generation is considered correct if it matches any variant of the requested Cipai.}
    \label{fig:variants_heatmap}
\end{figure*}

Comparing Figure~\ref{fig:direct_prompt_heatmap} and Figure~\ref{fig:variants_heatmap}, we observe that the heatmap colors for \textit{Cipai} forms such as ``金缕曲'' (Jin Lü Qu), ``贺新郎'' (He Xin Lang), ``临江仙'' (Lin Jiang Xian), ``南乡子'' (Nan Xiang Zi), and ``虞美人'' (Yu Mei Ren) have notably intensified. This indicates that under zero-shot instructions, models generated many valid variant forms for these particular \textit{Cipai}.
From Figure~\ref{fig:score_change_variants_num}, we can observe that when all variants are included in the evaluation, the average score increase for each \textit{Cipai} correlates with the number of variants it has.

\subsection{Cipai Name Confusion}
\label{app:name_confusion}

An inspection of Figure~\ref{fig:variants_heatmap} reveals a notable anomaly for the \textit{Cipai} ``浪淘沙'' (LangTaoSha). For this specific form, nearly all evaluated models failed to generate a structurally correct output according to our variant-aware evaluation. This discrepancy arises because ``浪淘沙'' and ``浪淘沙令'' (LangTaoSha Ling) are distinct \textit{Cipai} forms, despite their similar names. The standard ``浪淘沙'' consists of four seven-character lines, structurally identical to a seven-character quatrain (\textit{Qiyan Jueju}). In contrast, ``浪淘沙令'' is a longer form with 54 characters, more characteristic of typical \textit{Ci} poetry. Our analysis suggests that LLMs, when prompted with ``浪淘沙,'' predominantly generate the \textit{Cipai} ``浪淘沙令'' instead.

\begin{figure*}[p]
    \begin{tcolorbox}[width=\textwidth, colframe=blue!50!black, colback=blue!5!white, sharp corners, fontupper=\scriptsize]
        \textbf{Prompt:}\\
        请根据词牌``浪淘沙''创作一首词，主题为``春闺''。\\
        该词牌的格律要求如下：\\
        - 分句结构：中仄平平中仄平\quad 中平中仄仄平平\quad 中平中仄中平仄\quad 中仄平平仄仄平\\
        请直接输出词作，不需要解释。\\
        \textcolor{translationcolor}{Please compose a \textit{Ci} poem according to the \textit{Cipai} ``Lang Tao Sha'' and the theme ``Spring Boudoir''.\\
        Form requirements:\\
        - Sentence structure: neutral oblique level level neutral oblique level, neutral level neutral oblique oblique level level, neutral level neutral oblique neutral level oblique, neutral oblique level level oblique oblique level\\
        Please output the poem directly, without any explanation.}\\
        \textbf{Prediction:}\\
        深院晓莺惊梦残，\\
        画楼人静月空悬。\\
        红绡帐冷春寒透，\\
        独对菱花泪暗弹。\\
    \end{tcolorbox}
    \caption{A generation example using a form-aware prompt for 浪淘沙 (Lang Tao Sha). The confusion between ``浪淘沙'' and ``浪淘沙令'' is discussed in the main paper.}
    \label{fig:form_aware_langtaosha_example}
\end{figure*}

\subsection{Deepseek-R1 Reasoning Example}
\label{app:deepseek_example}

Figure~\ref{fig:deepseek_generation_example} shows a generation example from Deepseek-R1, including its detailed reasoning process. The model explicitly reasons about the \textit{Cipai} structure, tonal patterns, imagery selection, and rhyme scheme before generating the final poem.

\begin{figure*}[t]
    \begin{tcolorbox}[width=\textwidth, colframe=blue!50!black, colback=blue!5!white, sharp corners, fontupper=\scriptsize]
        \textbf{Instruction:} \\按照提供的词牌名和题目写一首词。要求严格遵守词牌的格律要求。\\
        \textcolor{translationcolor}{Write a \textit{Ci} poem according to the provided \textit{Cipai} name and theme. Strictly follow the metrical rules of the \textit{Cipai}.} \\
        \textbf{Input:}\\
        词牌: 浪淘沙 \\
        题目: 春闺 \\
        \textcolor{translationcolor}{Cipai: Lang Tao Sha \\
        Theme: Spring Boudoir} \\
        \textbf{Reasoning:}\\
        好的，用户让我按照提供的词牌名和题目写一首词，要求严格遵守格律。首先，我需要确认《浪淘沙》这个词牌的格律要求。《浪淘沙》通常有固定的格式，比如李煜的《浪淘沙令》是双调五十四字，上下片各五句四平韵。我得先回忆一下具体的平仄和押韵规则。\\
        \\
        接下来是题目《春闺》，主题应该是春天里女子的闺怨或思念。需要营造一种婉约、柔美的意境。可能需要用到一些传统意象，比如落花、帘幕、燕子、柳絮等，来表现春天的景象和内心的情感。\\
        \\
        然后要考虑结构，上下片如何分配内容。上片可能描绘春天的景色，下片转入情感的表达。要注意对仗和押韵的位置，确保符合《浪淘沙》的格律。比如每句的字数、平仄交替，以及押平声韵。\\
        \\
        \textcolor{translationcolor}{[The model's reasoning process: It first confirms the Cipai meter requirements, then considers the theme ``Spring Boudoir'' and appropriate imagery (fallen flowers, curtains, swallows, willow catkins). It plans the structure for upper and lower stanzas, checks tonal patterns and rhyme positions.]} \\
        \textbf{Output:}\\
        《浪淘沙·春闺》\\
        深院闭春光，落红飞过东墙。\\
        一桁水晶帘子底，燕子窥香。 \\
         \\
        残梦续新妆，瘦损潘郎。 \\
        闲掐青梅小朵簪，奁镜拂丝杨。何处是伊行，满径斜阳。\\
        \\
        注：我的创作遵循《浪淘沙》词牌双调五十四字之格律...\\
        \textcolor{translationcolor}{Lang Tao Sha: Spring Boudoir \\
        In deep courtyard, spring light confined, \\
        Fallen blossoms fly past the eastern wall. \\
        Beneath the crystal curtain on its rack, \\
        Swallows peek at fragrance...}
    \end{tcolorbox}
    \caption{A generation example of Deepseek-R1. It mixes two different Cipai, 浪淘沙 and 浪淘沙令.}
    \label{fig:deepseek_generation_example}
\end{figure*}

\section{Cipai Details for Non-Chinese Readers}
\label{app:cipai_details}

Table~\ref{tab:cipai_details_appendix} provides English translations and brief descriptions of the 30 \textit{Cipai} forms to aid non-Chinese readers.

\begin{table*}[htbp]
    \centering
    \caption{Details of the 30 \textit{Cipai} forms for non-Chinese readers.}
    \label{tab:cipai_details_appendix}
    \resizebox{\textwidth}{!}{%
    \begin{tabular}{@{}lll@{}}
    \toprule
    \textbf{Chinese Name (Pinyin)} & \textbf{English Translation} & \textbf{Brief Description} \\ \midrule
    浣溪沙 (Huàn Xī Shā) & Washing Creek Sand & Tang court music, 42 characters, three level-tone rhymes in first stanza, two in second \\
    鹧鸪天 (Zhè Gū Tiān) & Partridge Sky & Also called "Thinking of a Good Guest," 55 characters, three level-tone rhymes per stanza \\
    菩萨蛮 (Pú Sà Mán) & Bodhisattva Barbarian & Tang court music from foreign dance, 44 characters, alternating level and oblique rhymes \\
    蝶恋花 (Dié Liàn Huā) & Butterfly Loves Flower & Also called "Magpie Treads Branch," 60 characters, four oblique rhymes per stanza \\
    临江仙 (Lín Jiāng Xiān) & Riverside Immortal & Tang court music, 58 characters, three level-tone rhymes per stanza \\
    满江红 (Mǎn Jiāng Hóng) & Full River Red & 93 characters, passionate tone suitable for heroic emotions, typically uses entering-tone rhymes \\
    清平乐 (Qīng Píng Lè) & Clear and Peaceful Joy & Also called "Remembering Radish Moon," 46 characters, mixed rhyme scheme \\
    水调歌头 (Shuǐ Diào Gē Tóu) & Prelude to Water Melody & From Tang dynasty suite, 95 characters, four level-tone rhymes per stanza \\
    虞美人 (Yú Měi Rén) & Lady Yu & Named after Xiang Yu's concubine, 56 characters, alternating rhyme patterns \\
    沁园春 (Qìn Yuán Chūn) & Spring in Qin Garden & Grand format suitable for expressing magnificent emotions, 114 characters \\
    念奴娇 (Niàn Nú Jiāo) & Remembering Lady Niannu & Named after Tang dynasty singer, 100 characters, high-pitched heroic tone \\
    满庭芳 (Mǎn Tíng Fāng) & Courtyard Full of Fragrance & Also called "Locked Sun Terrace," 95 characters, level-tone rhymes \\
    西江月 (Xī Jiāng Yuè) & Moon over West River & Tang court music, 50 characters, mixed rhyme scheme with final oblique rhymes \\
    金缕曲 (Jīn Lǚ Qǔ) & Golden Thread Song & Same as "Congratulating the Bridegroom," 116 characters, six oblique rhymes per stanza \\
    点绛唇 (Diǎn Jiàng Chún) & Touching Rouge Lips & 41 characters, oblique rhymes, commonly used in opera \\
    减字木兰花 (Jiǎn Zì Mù Lán Huā) & Shortened Magnolia Flower & Reduced version of "Magnolia Flower," 44 characters, alternating rhymes \\
    踏莎行 (Tà Shā Xíng) & Treading on Grass & 58 characters, three oblique rhymes per stanza, opening with parallel couplets \\
    浪淘沙 (Làng Táo Shā) & Wave-Washed Sand & Originally seven-character quatrain, 28 characters, four level-tone rhymes \\
    水龙吟 (Shuǐ Lóng Yín) & Water Dragon Chant & Also called "Dragon's Song," 102 characters, four oblique rhymes per stanza \\
    望江南 (Wàng Jiāng Nán) & Gazing South of the River & Also called "Remembering Jiangnan," 27 characters, three level-tone rhymes \\
    如梦令 (Rú Mèng Lìng) & Like a Dream & Created by Later Tang Emperor Zhuangzong, 33 characters, five oblique rhymes \\
    南乡子 (Nán Xiāng Zǐ) & Southern Village Song & Tang court music, 27 characters in short form, 56 in long form \\
    贺新郎 (Hè Xīn Láng) & Congratulating the Bridegroom & Also called "Golden Thread Song," 116 characters, passionate tone \\
    卜算子 (Bǔ Suàn Zǐ) & Divination Song & Named after fortune tellers, 44 characters, two oblique rhymes per stanza \\
    采桑子 (Cǎi Sāng Zǐ) & Picking Mulberries & Also called "Ugly Slave," 44 characters, three level-tone rhymes per stanza \\
    摸鱼儿 (Mō Yú Ér) & Catching Fish & Also called "Buying Lotus Pond," 116 characters, complex rhyme structure \\
    忆江南 (Yì Jiāng Nán) & Remembering Jiangnan & Same as "Gazing South of the River," 27 characters, nostalgic theme \\
    渔家傲 (Yú Jiā Ào) & Fisherman's Pride & Northern Song popular song, 62 characters, five oblique rhymes per stanza \\
    江城子 (Jiāng Chéng Zǐ) & River Town Song & Also called "River God Song," 35 characters, five level-tone rhymes \\
    鹊桥仙 (Què Qiáo Xiān) & Immortal at the Magpie Bridge & Named after Qixi festival legend, 56 characters, about lovers' reunion \\
    \bottomrule
    \end{tabular}%
    }
\end{table*}

\section{Cipai Structure Examples}
\label{app:cipai_examples}

\begin{figure}[t]
    \centering
    \includegraphics[width=\columnwidth]{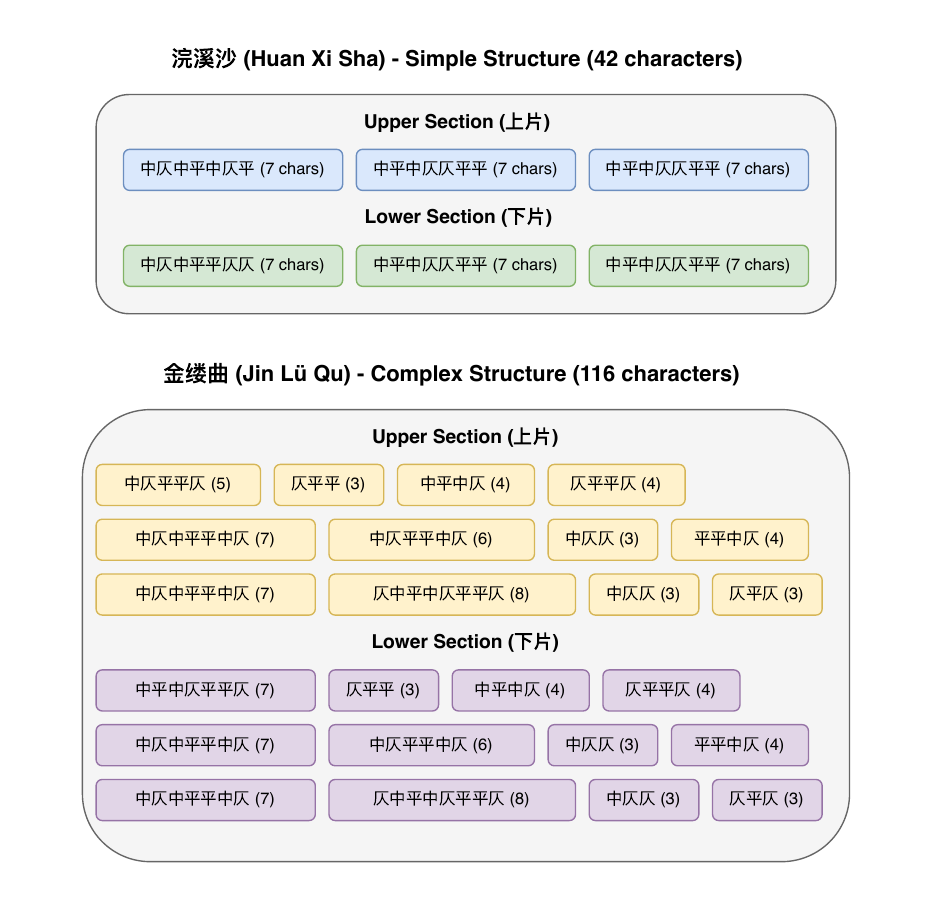}
    \caption{Examples of \textit{Cipai} structures: \textit{Huan Xi Sha} (simple, 42 characters) and \textit{Jin Lü Qu} (complex, 116 characters).}
    \label{fig:cipai_demo}
\end{figure}

Figure~\ref{fig:cipai_demo} visualizes two representative \textit{Cipai} forms. \textit{Huan Xi Sha} illustrates a short, regular structure (42 characters), while \textit{Jin Lü Qu} demonstrates a long, complex structure (116 characters) with more varied line-length patterns. This contrast highlights the wide range of structural difficulty covered by CCiV.

\section{Related Work}
\label{app:related_work}

Classical Chinese poetry generation has evolved from early approaches using template-based, statistical and human-involved methods~\cite{li2018GeneratingClassicalChinese,zhipeng2019JiugeHumanMachineCollaborative} that relied on limited keyword inputs and often produced rigid, semantically poor outputs~\cite{cao2024SurveyDeepLearning}. With the advancement of deep learning, recurrent neural networks (RNNs) and Transformer~\cite{vaswaniAttentionAllYou2017} models became mainstream approaches, offering better contextual understanding and overall consistency~\cite{wang2016ChinesePoetryGeneration,liu2020DeepPoetryChinese,shao2021SentimentStyleControllablea}. Yusheng~\cite{ma2023YuShengHumaninLoop} and Shisanbai~\cite{aichpoem2025} are two more practical poetry generation systems that based on generative pretrained transformers~\cite{radford2019LanguageModelsAre}. The emergence of large language models (LLMs) has brought new opportunities, allowing for unrestricted user prompts and enabling better control over content compared to traditional keyword-based systems~\cite{zou2024BIProZeroshotChinese}.

Despite these advances, most existing research has focused on the well-known \textit{Shi} forms, such as regulated verse (\textit{律诗}) and quatrains (\textit{绝句}). Although some benchmarks include tasks related to \textit{Ci} composition~\cite{cao2024WenMindComprehensiveBenchmark,zhao2025fuxibenchmarkevaluatinglanguage}, they are generally small in scale and lack detailed evaluation. The unique structural complexities and the challenge of numerous historical variants within \textit{Cipai} forms, which are central to our work, have remained largely underexplored.

\section{Limitations and Broader Impact}
\label{app:limitations}

Our study primarily focuses on structural and tonal accuracy and does not fully evaluate other critical dimensions of poetic quality, such as semantic coherence and overall artistic merit. We also do not evaluate rhyme positions or whether upper and lower stanzas share the same rhyme. Tonal labels are derived using \texttt{pypinyin} under Zhonghua Xinyun; polyphonic disambiguation is not handled, which may introduce noise. We see our work as enabling assistive tools that can make \textit{Ci} poetry more accessible, rather than replacing human creativity.

\noindent \textbf{Broader Impact and Ethical Considerations:}
Our goal is not to enable the mass production of poetry to replace human creativity. Instead, we envision our work as a foundational step toward developing assistive tools that can make this complex art form more accessible. \textit{Ci} poetry has a high barrier to entry, requiring deep literary and historical knowledge. An LLM-powered tool could act as a creative partner, helping aspiring poets and enthusiasts to learn the structural rules, experiment with different forms, and ultimately deepen their appreciation and engagement with this cultural heritage.

\end{document}